\documentclass[10pt,twocolumn,letterpaper]{article}

\usepackage{wacv}
\usepackage{times}
\usepackage{epsfig}
\usepackage{graphicx}
\usepackage{amsmath}
\usepackage{amssymb}

\wacvfinalcopy 

\def\wacvPaperID{****} 

\ifwacvfinal\fi
\begin{document}

\title{HistoNet: Predicting size histograms of object instances}



\author{Kishan Sharma \\
TU Munich\\
{\tt\small kishan.sharma@tum.de}
\and
Moritz Gold \\
ETH Zurich \& EAWAG\\
{\tt\small moritz.gold@hest.ethz.ch}
\and
Christian Zurbruegg\\
EAWAG\\
{\tt\small christian.zurbruegg@eawag.ch}
\and
Laura Leal-Taix\'e \\
TU Munich\\
{\tt\small leal.taixe@tum.de}
\and
Jan Dirk Wegner \\
ETH Zurich\\
{\tt\small jwegner@ethz.ch}}

\maketitle
\ifwacvfinal\thispagestyle{empty}\fi
\begin{abstract}
   We propose to predict histograms of object sizes in crowded scenes directly without any explicit object instance segmentation. What makes this task challenging is the high density of objects (of the same category), which makes instance identification hard. Instead of explicitly segmenting object instances, we show that directly learning histograms of object sizes improves accuracy while using drastically less parameters. This is very useful for application scenarios where explicit, pixel-accurate instance segmentation is not needed, but there lies interest in the overall distribution of instance sizes. Our core applications are in biology, where we estimate the size distribution of soldier fly larvae, and medicine, where we estimate the size distribution of cancer cells as an intermediate step to calculate the tumor cellularity score. Given an image with hundreds of small object instances, we output the total count and the size histogram. We also provide a new data set for this task, the FlyLarvae data set, which consists of 11,000 larvae instances labeled pixel-wise. Our method results in an overall improvement in the count and size distribution prediction as compared to state-of-the-art instance segmentation method Mask R-CNN~\cite{HeGDG17}.
\end{abstract}

\section{Introduction}
\noindent Pixel-wise segmentation of objects (e.g.,~\cite{shotton2006,gould2008,ronneberger2015,zhao2017}) and instance segmentation (e.g.,~\cite{bai2017,HeGDG17}) are core research topics in computer vision. Recent methods use Deep Neural Networks to estimate segmentation masks. Since the number of instances is not known a-priori, researchers either resort to object proposals first for detection~\cite{RenHG015} and later for segmentation~\cite{HeGDG17}, or enumerate instances with a Recurrent Neural Network \cite{romera2016recurrent}.
Both solutions require training of large models and complex training pipelines. 
In many applications, especially in the medical field, one is not interested in segmenting every instance of an object, but rather finding the distribution of object sizes in the image. 
Detecting object sizes in images is useful for a broad range of applications as it can be associated with physical properties like mass, area, etc. 
We target challenging tasks for which all objects have extremely similar appearance, and in which instance segmentation is challenging. Our first application is the prediction of size distributions of fly larvae colonies for organic waste decomposition~\cite{diener2011,gold2018}. Our second application aims at estimating tumor growth directly from medical images.

\noindent Typically, such tasks would be approached via explicit, pixel-accurate instance segmentation with a method like Mask R-CNN~\cite{HeGDG17}.
These methods can be used to predict the size of each individual object using the estimated mask. However, performance (for size estimation) decreases in case of overlapping objects and partial occlusions because only visible pixels can be classified and thus enter into the size estimation task. 
Furthermore, it is well known that instance segmentation methods cannot cope with large object overlap, mainly due to the Non-Maximum Suppression step, missing many objects in the process.
For those tasks, the size of instance segmentation models is disproportionate with respect to the complexity of the problem. 
In this paper, we advocate to learn and predict a statistical summary of object sizes and counts directly in the form of histograms. We show that our approach significantly reduces the parameter overhead needed for explicit, pixel-accurate instance segmentation, while being more accurate. Our {\bf contribution} is three-fold: 
\begin{itemize}
\item We propose a novel deep learning architecture (HistoNet), which counts and predicts the size distribution of objects directly from an input image, showing superior results with respect to state-of-the-art instance based segmentation methods while having 85\% less parameters.
\item We present a new data set\footnote{\url{ https://github.com/kishansharma3012/HistoNet}} of pixel-wise instance labeled fly larvae and the challenge of predicting size histograms for these small similarly-looking objects.
\item We further evaluate performance of HistoNet on a public cancer cell data set and demonstrate that it achieves good results for this different image modality and application domain.
\end{itemize}

\section{Related Work}
\noindent{\bf Counting and density estimation.}
Counting objects in images has been a focal point in computer vision research for several years. Various approaches of the pre-deep learning era designed bottom-up image processing workflows to count objects segmented with edge detectors~\cite{Edge1}. A downside of these approaches is their large number of hyper-parameters that has to be set for each new data set. Counting-by-regression methods~\cite{Neural1,Kong,Marana}, on the other hand, avoid directly solving the hard object detection problem but instead directly learn a mapping from the input image to the number of objects. 
An elegant method to estimate total object counts in images is density estimation. Lempitsky et al.~\cite{Lempitsky} densely computed Scale-Invariant Feature Transform (SIFT) features for the input images and predicted density maps via regression. Fiaschi et al.~\cite{Fiaschi} improved density mapping by using a regression forest.
Recent works turn to Deep Learning~\cite{rodriguez2018} for joint semantic segmentation and density estimation, to identify and count particular tree species of sub-pixel size at very large scale from satellite images. 
Xie et. al~\cite{Xie15} proposed a fully convolutional regression network to output a cell density map across the image to predict the total count. Similarly,~\cite{patch_based} used a structured regression convolutional neural network approach to detect cells.
A very powerful architecture custom-tailored for object counting, CountCeption was introduced by Cohen et. al~\cite{CohenLB17}. It processes the image in a fully convolutional manner and predicts redundant counting instead of density mapping in order to average over errors. Its main insight is that predicting the existence or absence of an object within a receptive field is an easier task than predicting density maps. The latter is harder because in addition to predicting an object's existence, it has to estimate how far the object is from the center of the receptive field. Due to its very redundant convolutions per image location, this architecture gives good results while being efficient to train. 
For many applications, object counting is not enough, as the size distribution of the objects in the scene is key to determine, e.g., malignant cell evolution. We propose an architecture that not only counts objects but also predicts their size distribution without explicit instance segmentation. 

\noindent{\bf Instance segmentation.}
An alternative way to predict total object count is to perform explicit object detection or instance segmentation, known as counting-by-detection. The last few years have seen considerable progress in object detectors~\cite{Girshick_RCNN,RenHG015}, as well as instance segmentation methods based on Deep Neural Networks~\cite{HeGDG17}. 
An advantage of these approaches is that they also provide object size as a by-product. This can be approximated by the area of the bounding box enclosing the objects or better estimated if one extracts instance masks for all objects.
To the best of our knowledge, this strategy is the most accurate and robust today to predict object count and size, thus we use it as the baseline for this work. 
A clear downside of the size and counting-by-detection strategy is that we solve a much harder problem, that of instance segmentation, in order to predict total counts and object size distributions. 
This means using large models and complex training schemes to obtain pixel-accurate instance delineation, even though this information is not needed as output.
Additionally, overlapping instances and occluded objects often lead to errors. 
Here, we propose to directly learn to predict size distributions without explicit instance segmentation or object detection, with higher accuracy and a leaner architecture having 85\% less parameters compared to Mask R-CNN~\cite{HeGDG17}.
\begin{figure*}
\begin{center}
   \includegraphics[width=0.65\linewidth]{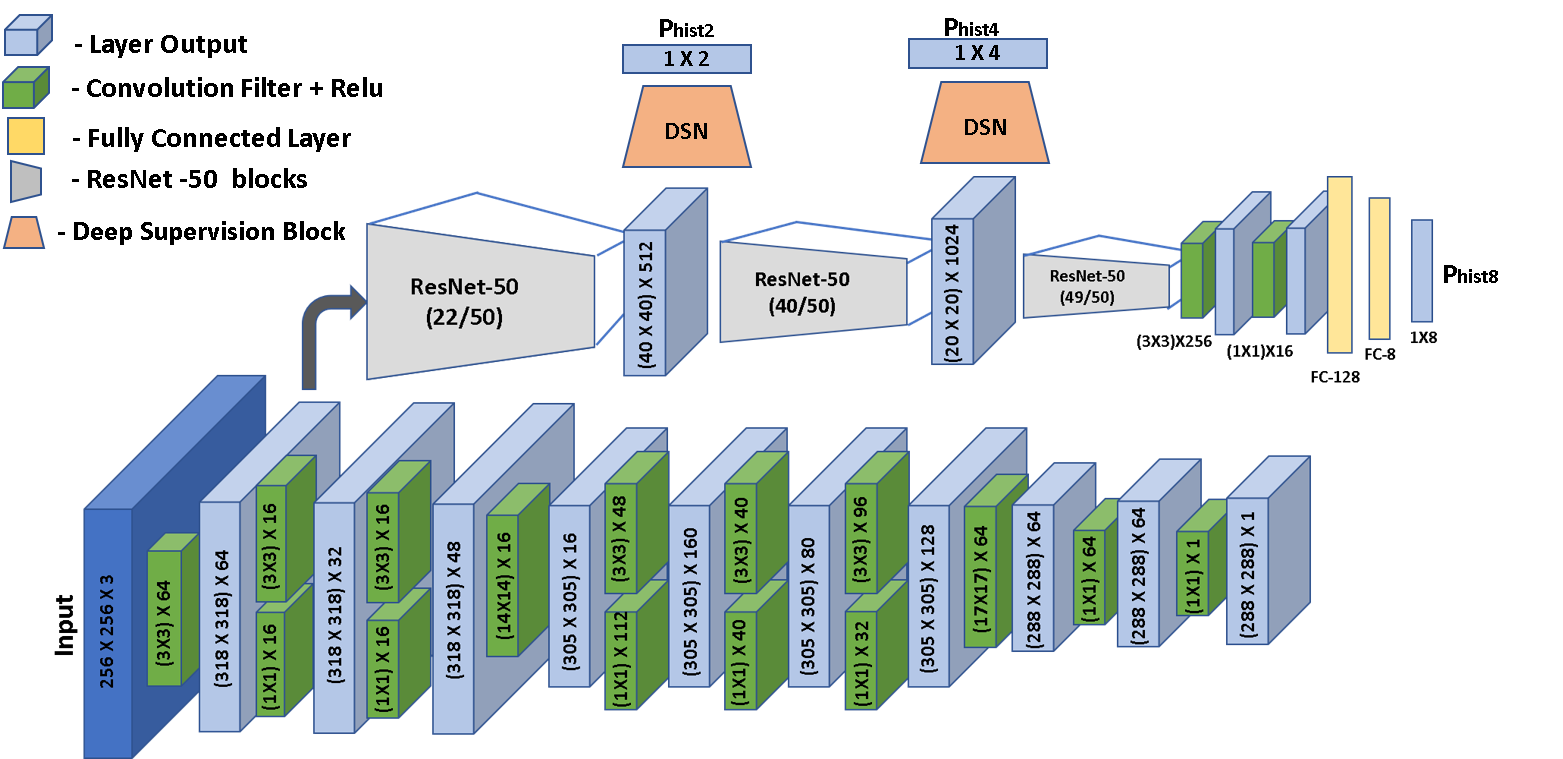}
\end{center}
\caption{HistoNet Deep Supervision Architecture}
\label{fig:histonet_dsn}
\end{figure*}
\section{HistoNet}
\noindent Our method directly predicts global data statistics given an image, namely, total object count in cluttered scenes and object size distribution via histogram prediction. 
More formally, given an input image $I$, our aim is to predict a redundant count map $P_{\text{map}}$ and size  histogram $P_{\text{hist}}$. 
Towards this end, we present a data-driven model, \emph{HistoNet}. Its architecture is shown in Fig.~\ref{fig:histonet_dsn}, while a notation summary for the following equations is given in Table~\ref{tab:notsum}.

\begin{table}
\begin{center}
\begin{tabular}{c||c}
\hline
Symbol & Description \\
\hline
I  & Input Image \\
$P_{\text{map}}$ & Predicted Count Map\\
$T_{\text{map}}$ & Target Count Map  \\
$P_{\text{hist}}$ & Predicted Histogram\\
$T_{\text{hist}}$ & Target Histogram\\
$p(H)$ & Probability Distribution of Histogram H\\
$L_{\text{count}}$ & Pixel Wise Count Map Loss \\
$L_{\text{wL}}$ & weighted L1 Histogram Loss \\
$W$ & Weights assigned for $L_{wL}$\\
$L_{\text{KL}}$ & KL-Div Histogram Loss \\
\hline
\end{tabular}
\end{center}
\caption{Notation Summary}
\label{tab:notsum}
\end{table}

\paragraph{Network Architecture.} 
As shown in Fig.~\ref{fig:histonet_dsn}, HistoNet consists of two branches, one for predicting the object count and the other for histogram prediction, which estimates the size distribution of object instances.
\emph{HistoNet} takes an image $I$ of size $256\times256$ as input and predicts a redundant count map $P_{\text{map}}$ of size $288\times288$ in a fully convolutional manner (lower branch). Note that neither upsampling nor deconvolutions are computed for predicting a map at a larger size than the input image. As in~\cite{CohenLB17}, \emph{HistoNet} zero-pads the input image in its first layer and we build a size distribution predictor network on top of it. Two green boxes in Fig.~\ref{fig:histonet_dsn} at the same level in the lower branch of the network represent the application of two kernels on the same input and concatenation of their outputs to handle variations in object size.
For size histogram prediction, the upper branch uses ResNet-50~\cite{HeZRS15} modules on top of the first layer of the lower branch.
Instead of a standard fully connected layer, we add two convolutional layers with kernels of size $3\times3\times256$ and $1\times1\times16$. These convolutional layers are followed by two fully connected layers interspersed with dropout layers. Our final output is a histogram of object sizes $P_{\text{hist}}$. 

\paragraph{Loss function.} To train  our multi-task prediction network, we impose losses on count prediction as well as on histogram distribution prediction. The count loss is L1 as in ~\cite{CohenLB17}:
\begin{equation}
L_{count}=  \left \| P_{map}(I) - T_{map}(I) \right \|_1
\end{equation}
where $P_{map}$ is the predicted count map, $T_{map}$ is target count map and $\left \| . \right \|_1$ is L1 norm.
\begin{figure*}
\begin{center}
\includegraphics[width=0.9\linewidth]{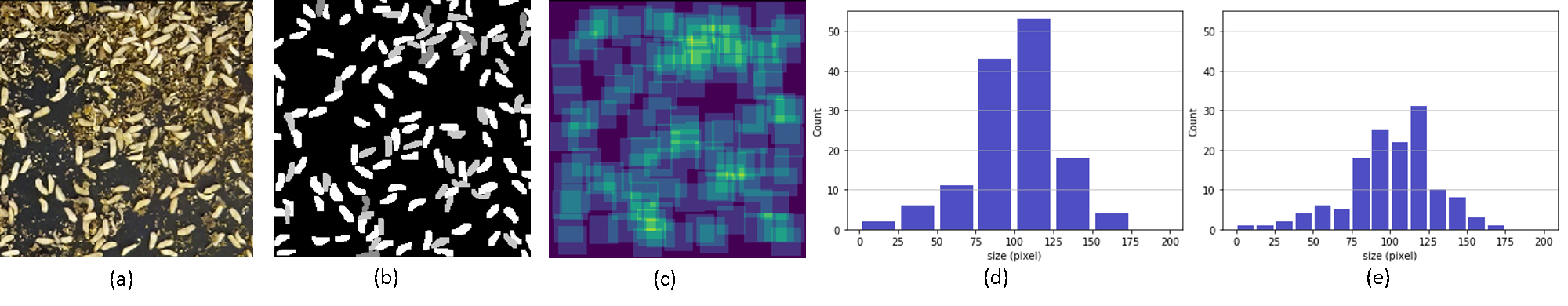}
\end{center}
   \caption{Example of our FlyLarvae data set: (a) input image, (b) pixel-accurate object masks, (c) count map (d) 8-bin histogram, (e) 16-bin histogram} 
\label{fig:Larvae_sample}
\end{figure*}

\begin{figure*}
\begin{center}
\includegraphics[width=0.8\linewidth]{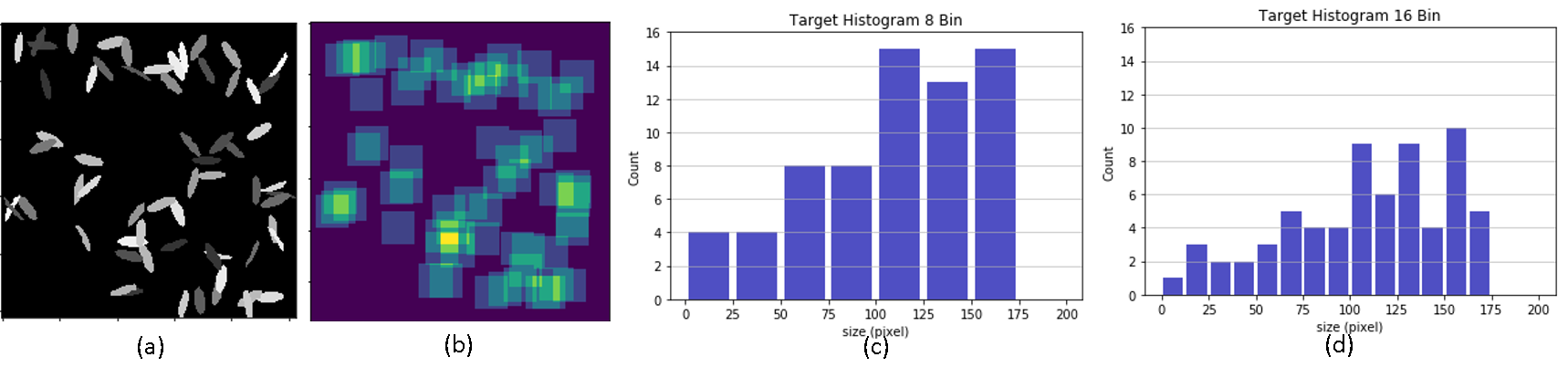}
\end{center}
   \caption{Example of the simulated ellipse data set: (a) input image (b) count map, (c) 8-bin histogram, (d) 16-bin histogram}
\label{fig:Synthetic_sample}
\end{figure*}
\begin{figure}

\begin{center}
\includegraphics[width=0.9\linewidth]{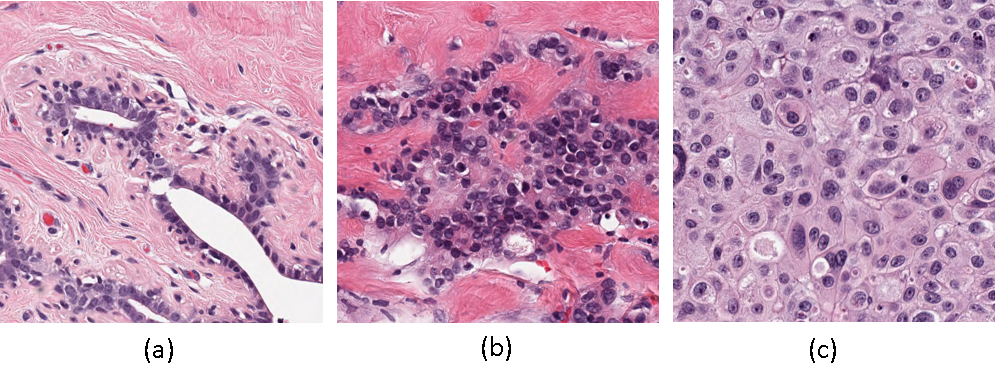}
\end{center}
   \caption{Example of breast cancer cell data set (a) Cellularity Score 0.0 (b) Cellularity Score 0.5 (c) Cellularity Score 1.0 }
  \label{fig:breast_cancer_cell}
\end{figure}
\noindent We formulate a KL-Divergence loss and a weighted L1 loss for size histogram prediction. The KL-Divergence loss measures the degree of dissimilarity between the predicted and ground truth distributions.
\begin{equation}
L_{KL}=  \sum p(T_{hist}) log\left ( \frac{p(T_{hist})}{p(P_{hist})}\right )
\end{equation}

Where $p(T_{hist})$ and $p(P_{hist})$ are the probability distributions of target and predicted size histogram. 
To capture the scale of the histogram, we found L1-loss to perform best. Moreover, a weighted L1-loss, where weights $W$ are assigned according to the normalized center values of the respective bins, further improves results. Our intuition is that larger objects should incur higher penalty than smaller ones if missed. For calculating weights, we (i) compute the bin centers, 
(ii) sum over the bin center vector, and (iii) divide with the total sum such that the overall sum across all bins is one

\begin{equation}
L_{wL}=  \sum W \left | P_{hist} - T_{hist} \right | 
\end{equation}
where $P_{hist}$ is predicted histogram and $T_{hist}$ is target histogram.
Our object size-weighted L1-loss in combination with the KL-Divergence loss are eventually mutually reinforcing to capture shape and scale of the histogram. We jointly train our network on this multi-task loss and minimize for count map and histogram prediction:
\begin{equation}
L_{total} =  L_{count} + 0.5 L_{KL} + 0.5 L_{wL}     
\end{equation}
We empirically found that giving equal weight to KL-loss and weighted L1-loss gives best results. 

\begin{figure*}
\begin{center}
\includegraphics[width=0.95\linewidth]{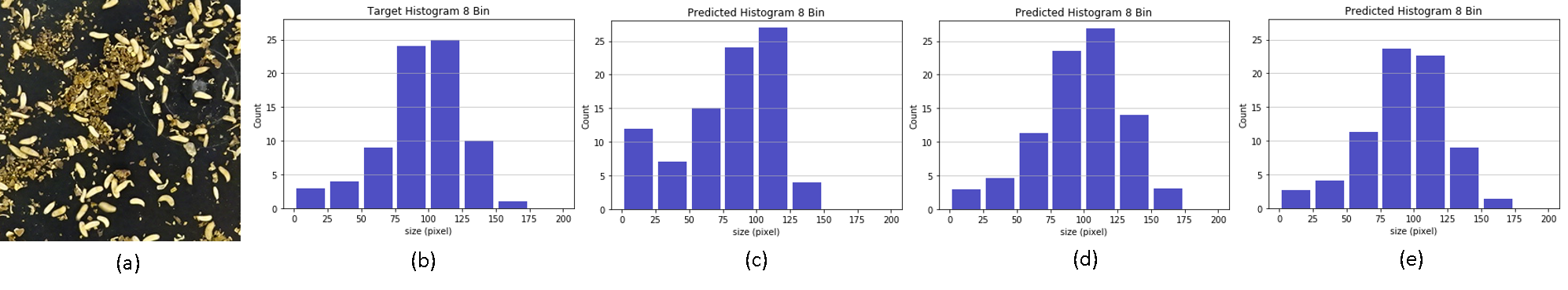}
\end{center}
   \caption{Fly Larvae 8-Bin Histogram Prediction (a) Input Image (b) Target 8-Bin Histogram (c) Mask R-CNN Prediction (d) HistoNet Prediction
   (e) HistoNet DSN prediction}
\label{fig:fly_larvae_pred}
\end{figure*}

\subsection{Deep supervision} 

\noindent Directly learning fine-grained 8 or 16 bin histograms can be tricky for the network. 
In order to help the network focus the learning on the hard cases near the bin boundaries, we propose to gradually increase histogram resolution towards the deeper layers. 
That is, we first learn a 2-bin and 4-bin histogram and later allow the network to increase the resolution to 8 and 16 bins without incurring any additional labelling cost. Towards this end, we use Deeply Supervised Nets (DSN)~\cite{LeeXGZT15}, shown to be helpful in calibrating the model at intermediate stages by enforcing direct and early supervision for both the hidden layers and the output layer. 
We show the Deep Supervision modules on the upper branch as trapezoids in Fig.~\ref{fig:histonet_dsn}. These have an effect on the early hidden layers and serve as an additional constraint 
to gradually force the network to split size bins into smaller intervals. Our deep supervision signal at early and middle stage of the histogram branch enforces first a split of sizes into two bins and the following one into four bins. 
Therefore, in addition to count map and size histogram, we predict 2-bin, 4-bin histograms $P_{\text{hist2}}$ and $P_{\text{hist4}}$, respectively. 
To implement deep supervision, we add a stack of two convolutions and two fully connected layers to predict the 2-bin histogram early in the histogram branch. The same stack is added at a later layer for the 4-bin histogram. Our full model architecture with deep supervision at intermediate stages of the histogram branch is shown in Fig.~\ref{fig:histonet_dsn}. 

For training the deeply supervised version of \emph{HistoNet} we define side output losses along side our main objective function. We add KL-divergence and weighted L1 loss for 2-bin and 4-bin histogram predictions: 
\begin{equation}
\begin{split}
L_{total} =  L_{count} + 0.5 ( L_{KL} + L_{wL} ) \\ + 0.2 ( L_{KL2} + L_{wL2} ) \\
+ 0.3( L_{KL4} + L_{wL4} )     
\end{split}
\end{equation}

\section{Experiments}
\noindent We evaluate performance of our method on three different data sets. \emph{HistoNet} performance is compared to state-of-the-art on our new \emph{FlyLarvae} data set, which contains a high density of similarly looking objects. In addition, we present experiments on a synthetic ellipse data set where we can adjust density and object size distributions arbitrarily, in order to show the robustness of our method and its ability to predict diverse histogram shapes. Finally, we run experiments with a medical image data set to verify applicability to a very different image modality and for the specific purpose of estimating the tumor cellularity score.

We use the Adam optimizer~\cite{kingma2014} for training our models and a batch size of four images. Our network weights are initialized using Xavier initialization~\cite{xavier}, and we apply classic data augmentation such as image vertical and horizontal rotation, noise addition, contrast variation and train our networks for 100 epochs.\\

\begin{figure*}
\begin{center}
\includegraphics[width=0.95\linewidth]{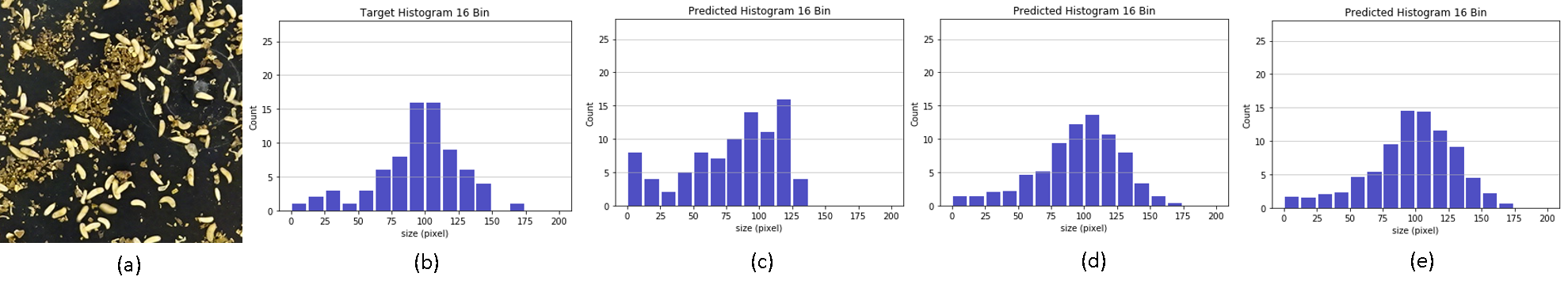}
\end{center}
   \caption{Fly Larvae 16 Bin Histogram prediction (a) Input Image (b) Target 16-Bin Histogram (c) Mask R-CNN Prediction (d) HistoNet Prediction (e) HistoNet DSN prediction}
\label{fig:fly_larvae_16_pred}
\end{figure*}

\begin{figure}

\begin{center}
\includegraphics[width=0.99\linewidth]{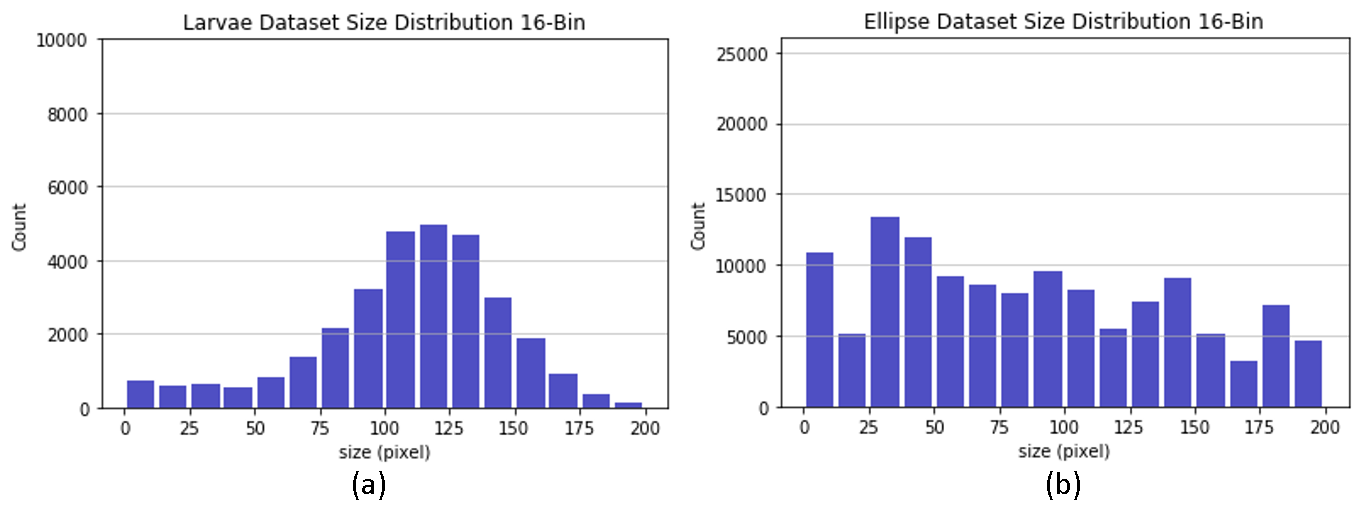}
\end{center}
\vspace{-0.2cm}
   \caption{Size distribution histogram (a) Fly larvae 16-bin (b) Synthetic ellipse 16-bin} 
\label{fig:dataset_size_distribution}
\end{figure}

\begin{figure*}
\begin{center}
\includegraphics[width=0.95\linewidth]{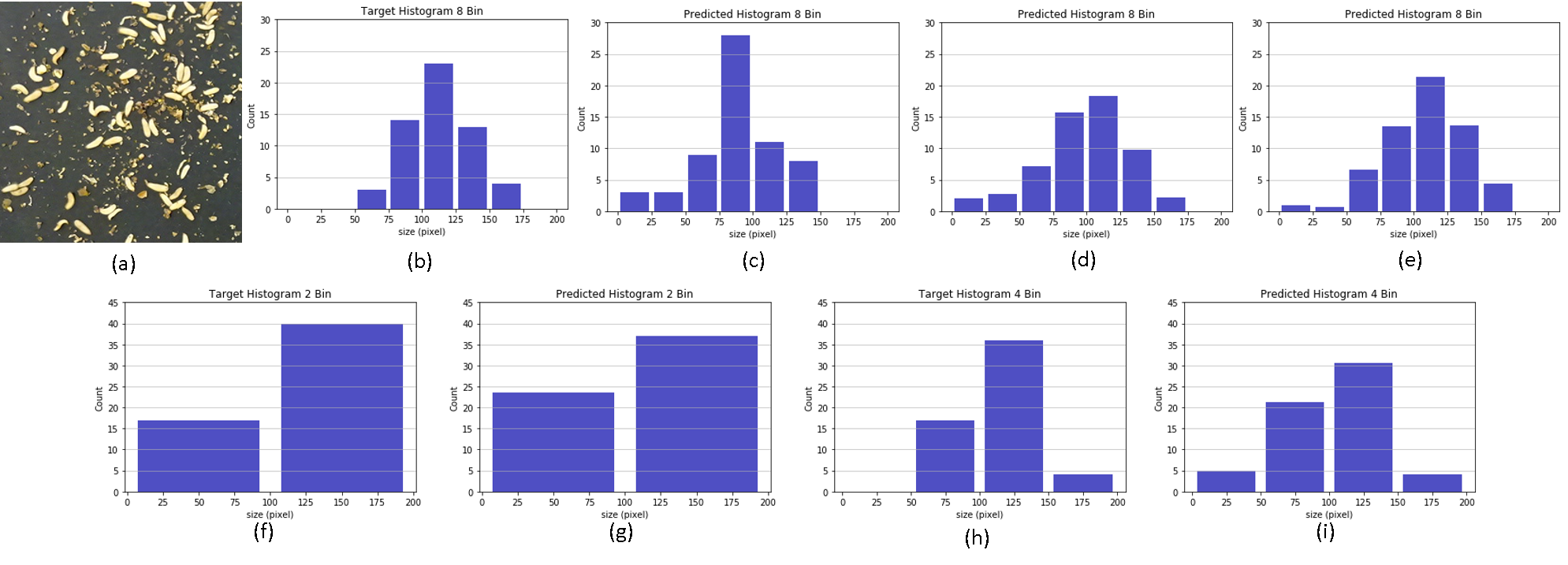}
\end{center}
   \caption{Fly Larvae 8 Bin Histogram prediction (a) Input Image (b) Target 8-Bin Histogram (c) Mask R-CNN Prediction (d) HistoNet Prediction (e) HistoNet DSN prediction (f) Target 2-Bin HistoNet DSN (g) Prediction 2-Bin HistoNet DSN (h) Target 4-Bin HistoNet DSN (i) Prediction 4-Bin HistoNet DSN}
\label{fig:fly_larvae_Dsn_pred}
\end{figure*}

\begin{figure*}
\begin{center}
\includegraphics[width=0.95\linewidth]{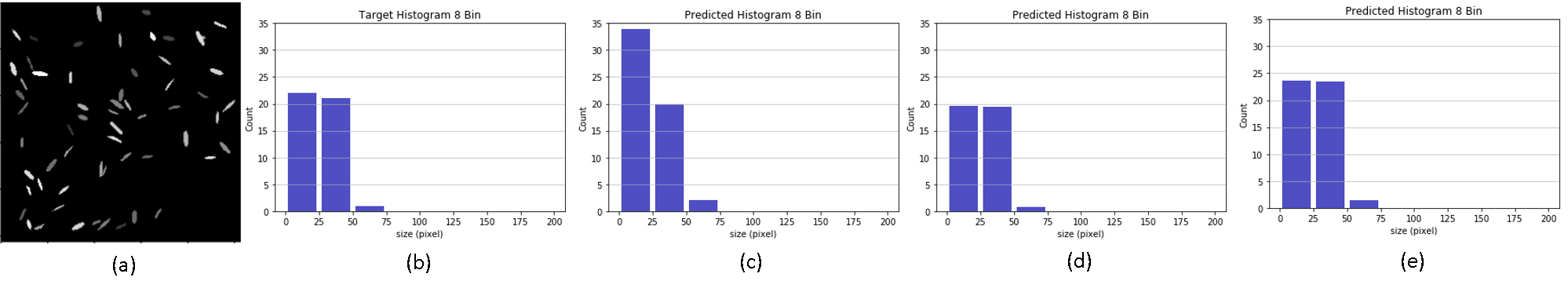}
\end{center}
   \caption{Synthetic Ellipse 8-Bin Histogram Prediction (a) Input Image (b) Target 8-Bin Histogram (c) Mask R-CNN Prediction (d) HistoNet Prediction (e) HistoNet-DSN Prediction}
\label{fig:Synthetic_pred8}
\end{figure*}

\begin{figure*}
\begin{center}
\includegraphics[width=0.95\linewidth]{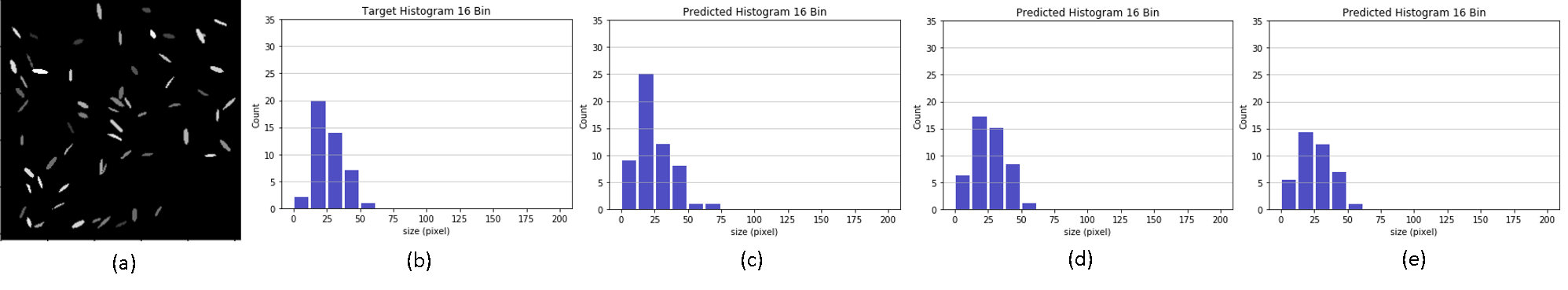}
\end{center}
\vspace{-0.2cm}
   \caption{Synthetic Ellipse 16 Bin Histogram prediction (a) Input Image (b) Target 8-Bin Histogram (c) Mask R-CNN Prediction (d) HistoNet Prediction (e) HistoNet-DSN Prediction}
\label{fig:Synthetic_pred16}
\end{figure*}

\noindent\textbf{Fly Larvae data set.}  We create a new dataset of soldier fly larvae, which are bred in massive quantities for sustainable, environmentally friendly organic waste decomposition~\cite{diener2011,gold2018}. Fly larvae images were collected using a Sony Cyber shot DSC -WX350 camera with image size $1380\times925$. The camera is installed on a professional repro tripod to guarantee a fixed distance from camera to object for all image acquisitions. Very large numbers of larvae mingled with a lot of larvae feed lead to high object overlap and occlusions. All larvae instances are labeled pixel-wise and we will make th \emph{FlyLarvae} dataset publicly available upon publication of this paper. 
For experiments, we sample patches of size $256\times256$ pixels from the original images. A summary of the \emph{FlyLarvae} data set is given in Tab.~\ref{tab:dataset_summary} while an example image and the corresponding instance, pixel-wise label is shown in Fig.~\ref{fig:Larvae_sample}. \\

\noindent\textbf{Synthetic Ellipse Dataset.} As evident from Fig.~\ref{fig:dataset_size_distribution}, the size distribution of our larvae data set follows a Gaussian distribution. In order to check whether our method really predicts different size distributions (or simply learns the Gaussian by heart), we created a synthetic data set of thin ellipses with strongly varying size distributions. We also greatly vary shape, size, and orientation of ellipses as well as the amount of overlap and occlusion. A summary of this synthetic ellipse data set is shown in Tab.~\ref{tab:dataset_summary} and an example image with its corresponding targets in Fig.~\ref{fig:Synthetic_sample}.\\
\begin{table}
\begin{center}
\begin{tabular}{|c|c|c|}
\hline
 & FlyLarvae & Synthetic Ellipse\\
\hline
No. Objects & 10844 & 135318\\
Size & 120.2 $\pm$ 28.1 & 94.5 $\pm$ 63.2  \\
Count & 80.4 $\pm$ 40.7 & 44.8 $\pm$ 20.8 \\
\hline
\end{tabular}
\end{center}
\caption{Summary of our new \emph{FlyLarvae} data set and the synthetic ellipse data set.}
\label{tab:dataset_summary}
\end{table}

\noindent\textbf{Breast Cancer Cell Dataset.} In order to validate applicability of \emph{HistoNet} to a different image modality and image content, we are using the breast cancer cell data set~\cite{BreastPathQData} that was originally recorded for the BreastPathQ Challenge.
It consists of 2579 image patches, and each patch is assigned a tumor cellularity score by one expert pathologist. The malignant cellularity score~\cite{BreastPathQ} depends on malignant cell count and size, a task which can be tackled with the proposed {\it HistoNet}. The BreastPathQ Challenge dataset~\cite{BreastPathQData} also contains a portion of images with annotated lymphocytes, malignant epithelial and normal epithelial cell nuclei. We add pixel-accurate labels to many of these images to prepare it for validating \emph{HistoNet}.  
Three example images of this data set are shown in Fig~\ref{fig:breast_cancer_cell}. 

\subsection{Evaluation Measures}
\noindent To evaluate object instance counts and size histogram prediction performance, we use several quantitative measures described in the following. For counting, we use the Mean Absolute Count Error (\emph{MAE}), which takes the absolute difference between predicted and target count.  
To quantify histogram prediction performance, we compute the Kullback-–Leibler divergence~\cite{kullback1951} (\emph{kld}), the Bhattacharyya distance~\cite{bhattacharyya1967} (\emph{bhatt}), the $\chi^{2}$-distance~\cite{nielsen2014}, intersection (\emph{isec}):
\begin{equation}
\begin{split}
isec = \frac{\sum \min(P_{hist},T_{hist})}{\max(\sum P_{hist}, \sum T_{hist} )}  
\end{split}
\end{equation}
and histogram correlation (\emph{corr}): 
\begin{equation}
\begin{split}
corr = \frac{\sum (P_{hist} - \overline{P_{hist}})(T_{hist} - \overline{T_{hist}})}{\sqrt{\sum (P_{hist} - \overline{P_{hist}})^{2}\sum (T_{hist} - \overline{T_{hist}})^{2}}}
\end{split}
\end{equation}
where $\overline{P_{hist}}$ and $\overline{T_{hist}}$ represent the mean of $P_{hist}$ and $P_{hist}$ histograms, respectively. 
\begin{table}
\begin{center}
\begin{tabular}{|c|c|}
\hline
 & Parameter number ($\times10^{6}$) \\
\hline
Mask R-CNN~\cite{HeGDG17} & 237.1\\
\emph{HistoNet} & 30.2  \\
\emph{HistoNet DSN} & 36.5 \\
\hline
\end{tabular}
\end{center}
\caption{Total parameter number per model.}
\label{tab:Parameter}
\end{table}

\begin{table*}[ht]
\centering
\begin{tabular}{p{3.2cm} || p{1.3cm} p{1.3cm} p{1.3cm} p{1.8cm} p{1.3cm} p{1.8cm} p{1.5cm}}
\hline
Method & MAE $\downarrow$ & \emph{kld} $\downarrow$ & $wt_{\text{L1}}$ $\downarrow$ & \emph{isec} $\uparrow$ & $\chi^{\text{2}}$ $\downarrow$ & \emph{corr} $\uparrow$ & \emph{bhatt} $\downarrow$\\
 
\hline
Average model 8 & 28.97 & \textbf{0.10} & 4.42 & 0.66 & 13.24 & 0.90 & 0.17\\
Mask R-CNN 8 & 7.84 & 0.64 & 4.31 & 0.72 & 16.37 & 0.77 & 0.25 \\
HistoNet 8 & 2.38 & 0.25 & 2.72 & 0.81 & 6.57 & 0.91 & 0.16\\
HistoNet-DSN 8 & \textbf{2.06} & 0.23 & \textbf{2.51} & \textbf{0.83} & \textbf{5.74} & \textbf{0.93} & \textbf{0.15}\\
\hline
Average model 16 & 28.97 & \textbf{0.17} & 2.46 & 0.64 & 17.12 & 0.84 & 0.22\\
Mask R-CNN 16 & 7.84 & 0.95 & 2.62 & 0.69 & 22.73 & 0.69 & 0.32 \\
HistoNet 16 & 2.28 & 0.26 & 1.74 & 0.76 & 10.03 & \textbf{0.86} & \textbf{0.21} \\
HistoNet-DSN 16 & \textbf{1.99} & 0.25 & \textbf{1.70} & \textbf{0.77} & \textbf{9.8} & \textbf{0.86} & \textbf{0.21}\\
\hline
\end{tabular}
\caption{FlyLarvae data set}
\label{tab:flylarvae}
\end{table*}

\begin{table*}[ht]
\centering
\begin{tabular}{p{3.2cm}||p{1.5cm}p{1.5cm}p{1.5cm}p{1.5cm}p{1.5cm}p{1.5cm}p{1.5cm}}
\hline
Method & MAE $\downarrow$ & \emph{kld} $\downarrow$ & $wt_{\text{L1}}$ $\downarrow$ & \emph{isec} $\uparrow$ & $\chi^{\text{2}}$ $\downarrow$ & \emph{corr} $\uparrow$ & \emph{bhatt} $\downarrow$\\
\hline
Average model 8 & 15.74 & 0.58 & 3.52 & 0.46 & 21.2 & 0.26 & 0.43\\
Mask R-CNN 8& 4.02 & 0.50 & 1.67 & 0.75 & 6.01 & 0.75 & 0.22 \\
HistoNet 8 & 1.64 & 0.17 & 1.22 & 0.73 & 3.81 &0.82 & 0.17\\ 
HistoNet-DSN 8 & \textbf{1.2} & \textbf{0.13} & \textbf{1.17} & \textbf{0.78} & \textbf{3.36} & \textbf{0.83} & \textbf{0.16}\\
\hline
Average model 16 & 15.74 & 0.70 & 1.89 & 0.43 & 24.03 & 0.27 & 0.48\\
Mask R-CNN 16 & 4.02 & 1.12 & 1.13 & 0.67 & 10.81 & 0.64 & 0.32 \\
HistoNet 16 & 1.45 & 0.47 & 0.87 & 0.68 & 7.19 & 0.71 & 0.27\\
HistoNet-DSN 16 & \textbf{1.09} & \textbf{0.24} & \textbf{0.85} & \textbf{0.69} & \textbf{6.70} & \textbf{0.74} & \textbf{0.25}\\
\hline
\end{tabular}
\caption{Synthetic Ellipse data set}
\label{tab:synthetic_Result}
\end{table*}

\subsection{Quantitative Results on biological data}
\noindent We evaluate \emph{HistoNet} for predicting 8-bin histograms and more fine-grained 16-bin histograms of object sizes. We benchmark \emph{HistoNet} against Mask R-CNN~\cite{HeGDG17} as a baseline. Recall that Mask R-CNN predicts pixel-accurate instance labels instead of directly outputting an object size distribution. We thus explicitly do instance segmentation and compute sizes by summing over instance pixels. We compare performance of explicit instance segmentation~\cite{HeGDG17}, our \emph{HistoNet} and \emph{HistoNet DSN} architectures on \emph{FlyLarvae} (Tab.~\ref{tab:flylarvae}) and the synthetic ellipse data set (Tab.~\ref{tab:synthetic_Result}).\\ 

\noindent{\bf FlyLarvae Dataset. }For the \emph{FlyLarvae} data set, our approach reduces the  $\chi^{2}$-distance for histogram prediction by more than 50\% compared to the Mask R-CNN baseline. In addition, significantly improved Kullback--Leibler divergence (\emph{kld}) and weighted L1 difference between histograms ($wt_{L1}$) indicate that our method captures scale and shape of the histograms much better than Mask R-CNN. As shown in Fig.~\ref{fig:fly_larvae_pred},~\ref{fig:fly_larvae_16_pred} our approach predicts histograms which are close to the ground truth size histograms. Mask R-CNN over and under-predicts the masks of objects, and thus their size.  Fig.~\ref{fig:fly_larvae_Dsn_pred} shows the histogram prediction provided by the deep supervision in the form of 2-bin and 4-bin histogram demonstrating that \emph{HistoNet DSN} further improves over \emph{HistoNet} method.\\

\begin{table}[hbt]
\begin{center}
\begin{tabular}{|c|c|}
\hline
 Method & Prediction Probability \\
\hline
CountCeption & 0.56\\
HistoNet-[fc 128, fc 128] & 0.69\\ 
HistoNet-[fc 18, fc 18] & 0.76 \\
HistoNet-[fc 32, fc 32] & 0.79 \\
HistoNet-[fc 64, fc 64]  & \textbf{0.83} \\
\hline
\end{tabular}
\end{center}
\caption{Breast Cancer Cell - Cellularity score prediction}
\label{tab:breast_cancer_result}
\end{table}

\vspace{-0.2cm}
\noindent{\bf Synthetic Ellipse Dataset.} Since the {\it FlyLarvae} dataset approximately follows gaussian distribution, one could assume that a model that predicts the average shape of the training set would be able to solve the task. In order to show that this is not the case, we create a synthetic ellipse dataset which covers a much higher variance of size distributions, density, and object overlaps. As depicted in the Table~\ref{tab:synthetic_Result}, similar trends are observed on synthetic ellipse data set for $\chi^{2}$-distance and correlation between histograms. Our method is able to handle high variance in object sizes and thus showing robustness on synthetic ellipse data set.
 We clearly show that even if the histogram is skewed, {\it HistoNet} is able to correctly predict its shape. We generalize to the number of objects, their size and well as their distribution of sizes. Note, that our method uses 85\% less parameter as compared to Mask-RCNN as shown in Tab.~\ref{tab:Parameter}. \\

\vspace{-0.2cm}

\noindent{\bf Average model.} To further test whether our model is just learning the average distribution shape of the training set, we compare to a baseline \emph{Average Model} that does exactly that: it computes the average object count and size histogram of the training data and uses that as predictions for the test set.
%
%
Since the FlyLarvae dataset approximately follows a Gaussian distribution, \emph{Average Model} is able to capture the probability distribution of the histogram, as shown by the low \emph{kld} in Tab.~\ref{tab:flylarvae}. Nonetheless, \emph{Average Model} performs very poorly on all other measures, demonstrating that our \emph{HistoNet} model can go beyond averaging training data distribution and count. {\it HistoNet} correctly predicts the shape and scale of the histograms.
As expected, Tab.~\ref{tab:synthetic_Result} shows that \emph{Average Model} fails to capture the underlying distribution and scale of the synthetic ellipse dataset histogram, which has a much larger variance in object sizes and a large variety of object size histograms.

\subsection{Quantitative Results on medical data}
\noindent The Breast Cancer Cell Dataset~\cite{BreastPathQData} depicts the cellularity score of a patch, which directly depends on the area of malignant cells. For this task, we first train our \emph{HistoNet} model to predict the count map and size distribution histogram of malignant cells. 
Using the countmap and histogram prediction, we fine-tune our model to predict the cellularity score. The whole network is end-to-end trainable, but due to lack of pixel-wise labeled data, we use a multi-part training schedule to train this network. 
\begin{itemize}
    \item \textbf{Stage 1:} Using the additional dataset in BreastPathQ challenge, for which nuclei information is provided, we created our target redundant count map. So, we trained only the lower branch of HistoNet to predict the redundant count map. 
    \item \textbf{Stage 2:} We manually pixel-wise labeled some images from the additional dataset, to train the upper branch of HistoNet, to learn to predict 8-bin size histogram. On this small dataset, we train the whole HistoNet. 
    \item \textbf{Stage 3}: We use the main dataset, which has images labeled with their cellularity score, to train the remaining part of this architecture. During this training, we fix the weights learned from stage-2 for HistoNet.   
\end{itemize}

We compare our method with a CountCeption-based cellularity score prediction model. To evaluate our method for cellularity score prediction, we follow the challenge rules, and use the prediction probability measure. This is calculated for each method for each reference standard (pathologist 1 and pathologist 2), then averaged to determine a final overall prediction probability value. 
Our method significantly improves the cellularity score prediction over the CountCeption-based method. This indicates that merely information about the malignant cell count is not sufficient to predict the cellularity score with good accuracy. Estimating the size distribution histogram significantly helps and improves prediction accuracy. Among the variants of \emph{HistoNet} we found using two fully connected layers of size 64 to predict cellularity score performs best as shown in Tab.~\ref{tab:breast_cancer_result}.\\

\begin{figure}[t]
\begin{center}
\includegraphics[width=0.99\linewidth]{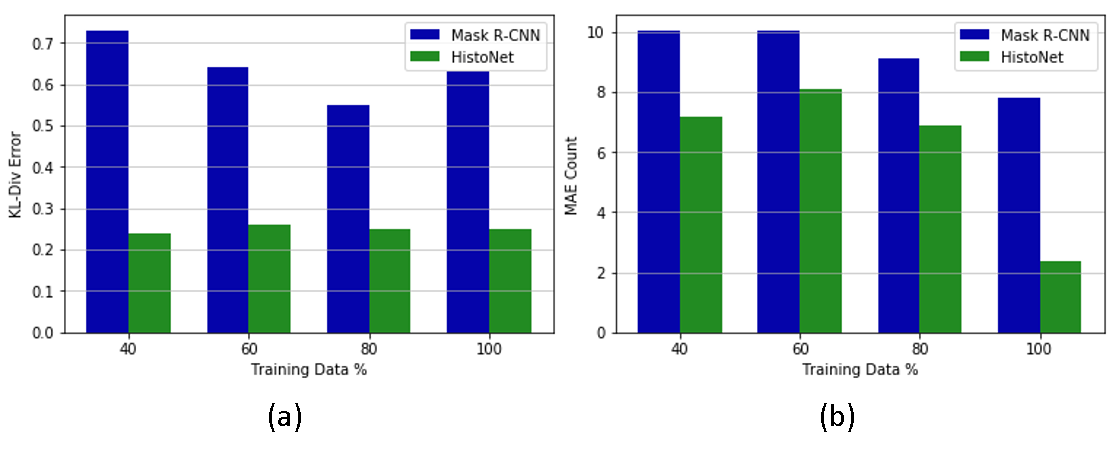}
\end{center}
   \caption{Ablation Study (a) KL-Divergence Error (b) Mean absolute count error}
\label{fig:ablation}
\end{figure}

\noindent\textbf{Ablation Study.} We finally perform an ablation study to evaluate how the performance of the method changes with the amount of training data. Because of the scarcity of labeled data in biomedical applications, it is important to design methods that are lightweight and can be trained without resorting to a large number of labeled examples.
As we show in Fig.~\ref{fig:ablation}, as the amount of training data is reduced, Mask-RCNN results for \emph{kld} and \emph{MAE} error increase, while our approach requires less training data to achieve better results.
\section{Conclusion}
\noindent We have presented \emph{HistoNet}, a new deep learning approach that predicts object size distributions and total counts in cluttered scenes directly from an input image. Experimental evaluation on a new FlyLarvae data set and a medical data set show superior performance compared to explicit object instance segmentation methods and data-driven methods that predict only object counts.
We verify with synthetic images of strongly varying object densities and object overlap that our method can predict a diverse set of size histogram shapes. 
We show that directly learning and predicting object size distributions, without a detour via explicit, pixel-accurate instance segmentation, significantly improves performance. In addition, we save 85\% of model parameters, which leads to a much leaner architecture that can be trained with fewer annotations.
We believe that the value of direct histogram prediction goes beyond our specific use cases. In future work, we will investigate its potential to significantly speed up state-of-the-art object detectors by modelling spatial priors on anchor box distributions, which is mostly done in a greedy fashion nowadays.  
\paragraph{Acknowledgement:} The BreastPathQ Challenge data used in this research was acquired from Sunnybrook Health Sciences Centre with funding from the Canadian Cancer Society and was made available for the BreastPathQ challenge, sponsored by the SPIE, NCI/NIH, AAPM, Sunnybrook Research Institute.

{\small
\bibliographystyle{ieee}
\bibliography{egpaper_final}
}

\end{document}